\documentclass[letterpaper, 10 pt, conference]{ieeeconf}  

\IEEEoverridecommandlockouts                              

\usepackage{caption}

\usepackage{graphicx}
\usepackage{comment}
\usepackage{booktabs}
\usepackage{amsmath,amssymb} 
\usepackage{color}
\usepackage{gensymb}
\usepackage{subfigure}
\usepackage{stackengine}
\usepackage{multirow}
\usepackage{color}

\usepackage[table]{xcolor}
\usepackage{floatrow}
\usepackage{bm}
\newcommand{\cmmnt}[1]{}

\title{\LARGE \bf
SSP: Single Shot Future Trajectory Prediction
}

\author{Isht Dwivedi \and Srikanth Malla \and Behzad Dariush \and Chiho Choi
\thanks{The authors are with Honda Research Institute USA, 70 Rio Robles, San Jose, CA 95134, USA.
{\tt\small \{idwivedi,smalla, bdariush,cchoi\}@honda-ri.com}}
}

\begin{document}

\maketitle
\thispagestyle{empty}
\pagestyle{empty}

\begin{abstract}
We propose a robust solution to future trajectory forecast, which can be practically applicable to autonomous agents in highly crowded environments. For this, three aspects are particularly addressed in this paper. First, we use composite fields to predict future locations of all road agents in a single-shot, which results in a constant time complexity, regardless of the number of agents in the scene. Second, interactions between agents are modeled as a non-local response, enabling spatial relationships between different locations to be captured temporally as well (i.e., in spatio-temporal interactions). Third, the semantic context of the scene are modeled and take into account the environmental constraints that potentially influence the future motion. To this end, we validate the robustness of the proposed approach using the ETH, UCY, and SDD datasets and highlight its practical functionality compared to the current state-of-the-art methods. 
\end{abstract}

\section{INTRODUCTION}
Predicting future motion of agents from video inputs in indoor and outdoor environments is a fundamental building block in various applications such as autonomous navigation in robotics, autonomous driving and driving assistance technologies, and video surveillance. Although substantial progress has been made in trajectory forecast, existing approaches have not fully addressed important challenges as follows: (i) interactions between agents are modeled without proper consideration of how the agents influence one another in both spatial and temporal domains, (ii) surrounding environmental constraints are to a large extent ignored, and (iii) perhaps most importantly, the computational complexity of existing methods linearly increases with respect to the number of agents, which places strict limits on their practical use in real-time and safety critical applications such as autonomous navigation for robotics or driving scenarios.

In this work, we provide a robust and computationally efficient solution to address the aforementioned challenges. Our method primarily aims for single-shot prediction of the trajectory of all agents in the scene. This is undoubtedly important for applications involving intelligent mobility (\textit{e.g.}, automated guide/service robots, self-driving vehicles, and driving assistance technologies) where the trajectory prediction is safety critical and has real time requirements using minimal compute resources. To achieve single shot prediction, the proposed method makes use of two types of composite fields: (i) a \textit{localization field} that predicts the future locations of all agents at each time step and (ii) an \textit{association field} to associate the predicted locations between successive frames. By associating composite fields from the last observation, identification of each agent directly propagates through future time steps. To the best of our knowledge, this work is the first to predict all agents' future trajectories in a single-shot manner, running in constant time complexity.

Our framework also models interactions with the surrounding environment,  including interactions with other agents in the scene as well as physical constraints such as obstacles or structures. It should be noted that the convolutional and recurrent operations in many existing approaches capture local interactions, either in space~\cite{fukushima1982neocognitron,lecun1989backpropagation} or in time~\cite{rumelhart1986learning,hochreiter1997long}, but not both. Thus, the interaction modules of the existing trajectory prediction approaches~\cite{alahi2016social,gupta2018social,cui2019multimodal} have inherent limitations in capturing both spatial and temporal relationships. However, recent work on a non-local block~\cite{wang2018non} have shown promising results in modeling such spatio-temporal relationships (\textit{i.e.}, in space-time). The non-local block computes the features at a location as a linear weighted combination of features at all locations in space and in time. 
Inspired by their non-local operator functions, we model social interactions of entities in space-time domains and further extend the function of non-locality to capture the environmental interaction by constraining the operation using the semantic context of the scene. In this way, our new proposed block is able to capture both the inter-agent and environmental interactions.

\begin{figure}[!tbp]
\centering
\includegraphics[width=1 \textwidth]{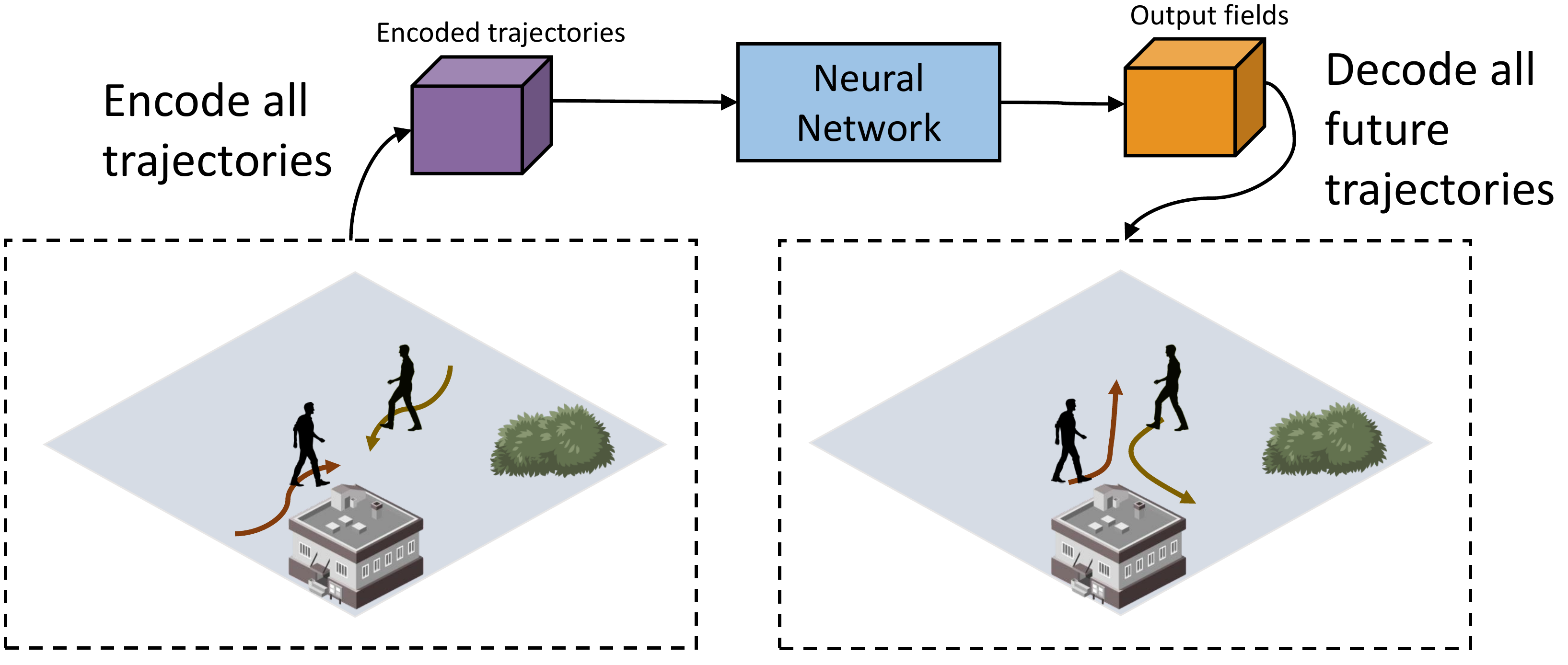}
\caption{Our network runs in constant time with respect to the number of pedestrians: past trajectories of all agents are encoded in a map (shown by the purple block), sent to a neural network, which produces output fields (shown by the orange block) in $1$ forward pass. The output fields are decoded to produce future trajectories of all agents.}
\label{fig:main}
\end{figure}

The main contributions of this work are as follows:
\begin{enumerate}
    \item The use of composite fields enables us to predict the future trajectory of all road agents with one-shot forward pass computation. Thus, the network run time is constant, O(1),  with respect to any number of agents in the scene.
    \item The interactive behavior of entities is jointly captured both spatially and temporally, which models the social interaction between agents in the scene.
    \item The proposed non-local module integrates visual semantics of the physical environment to take into account spatio-temporal interactions with the physical environment.
\end{enumerate}

\section{Related Work}
In this section, we review deep learning-based literature on trajectory prediction, non-local interaction, and convolution LSTMs.

\noindent
\textbf{Trajectory Prediction.} Early methods focused on modeling social interactions between humans. The social pooling layer is introduced in \cite{alahi2016social} and extended in \cite{gupta2018social}. Both methods use the recurrent operation for individual pedestrians, which aims to capture their temporal relationships. Although they motivated subsequent research such as \cite{lee2017desire,sadeghian2019sophie,sadeghian2018car,yao2019egocentric,malla2019nemo,ivanovic2019trajectron,zhang2019sr,malla2020titan} to process neighborhoods agents, spatial relationships are not explicitly modeled in space. Another research thread~\cite{yagi2018future,nikhil2018convolutional,su2019potential} uses the convolution operation to encode past motion of pedestrians as well as their interactions. However, the fact that these methods do not consider temporal changes makes them susceptible to errors in capturing long-range interactions. The recent work~\cite{choi2019looking,choi2020shared} considers both spatial and temporal interactions of agents. However, their successive 2D-3D convolutional operations are computationally inefficient. 


\noindent
\textbf{Convolutional LSTM.} 
Traditional LSTM has been extended to Convolutional LSTM (Conv-LSTM) for the direct use of images as input for precipitation nowcasting \cite{xingjian2015convolutional}. 
Subsequently, \cite{finn2016unsupervised} proposed an architecture consisting of several stacks of Conv-LSTM for learning physical interaction through video prediction. The success of such works inspired us to encode past motion of all road agents using images through Conv-LSTM layers.
Unlike existing methods that encode a single motion history, we encode the features using a set of binary maps. Each binary map indicates the locations of all road agents at a certain time step. 

\noindent
\textbf{Single-shot computation.} Estimation of multiple locations in a single-shot manner has been recently studied. In pose estimation, the joint locations of human body and their connections were estimated using a single-shot framework \cite{papandreou2018personlab, kreiss2019pifpaf}. In parallel, an encoder-decoder architecture was used to synthesize future motion sequences producing a set of motion flow fields in \cite{ji2018dynamic}. Motivated by these works, we propose a framework to predict future trajectories of all road agents in single-shot. The use of composite fields (\textit{i.e.}, localization and association field) enables us to locate the future positions of all agents at each time step and to find their associations between successive frames.



\noindent
\textbf{Non-local interaction.} In contrast to existing methods in trajectory prediction, we model interactions between agents as a non-local response in space-time domains. Non-local algorithms were introduced in \cite{buades2005non} for image de-noising. This idea was recently adapted to video action recognition in \cite{wang2018non} for neural networks. \cite{hussein2019timeception} subsequently used multi-scale temporal convolutions for long-range temporal modeling \cmmnt{non-local interaction} in videos. For future trajectory prediction applications, we adapt the idea in \cite{wang2018non} that was originally proposed for action recognition. Our framework captures not only the spatio-temporal interactions between agents, but also the interaction with the environment using a novel interaction module, exhibiting  improved performance in our experiments.

\section{Methodology}
\begin{figure*}[!t]
\centering
 \includegraphics[width=0.98\textwidth,keepaspectratio]{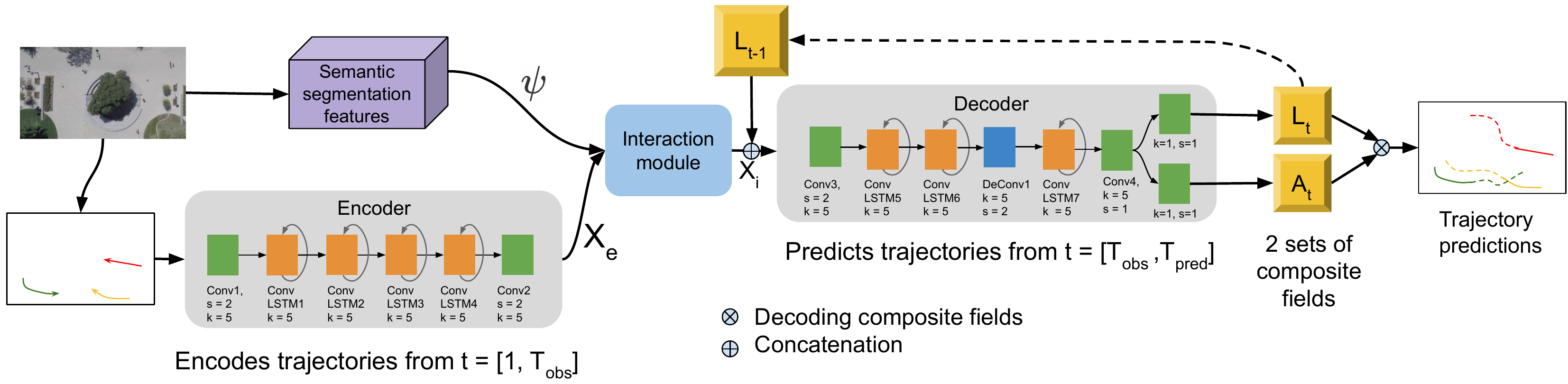}
\caption{Network Architecture showing the encoder, decoder, interaction modules and the produced composite maps. The composite maps $L_t, A_t$ are decoded to produce output trajectories. The parameters s and k refer to stride and kernel size, respectively. Padding for each layer is selected such that the output remains a multiple of 32.}
\label{fig:1}
\end{figure*}
We observe the trajectories of all agents for $T_{obs}$ time steps. The position of pedestrian $i$ at time $t$ is denoted by $P_i^{t}=(x_i^{t},y_i^{t})$. We predict the trajectories of all agents for $T_{pred}$ time steps in the form of two composite fields  $L_t$ and $A_t$, representing \textit{localization} and \textit{association} fields, respectively. 

\subsection{Network Architecture}
Our trajectory prediction network consists of three sub-modules: past motion encoder, interaction module, and future motion decoder as shown in Fig.~\ref{fig:1}. The past motion encoder and future motion decoder consist of a set of convolutional and Conv-LSTM layers. We generate an image-like tensor using the positions of all pedestrians at each past time step, which is used as input to the past motion encoder. 
The output encoding $X_e$ of the past motion encoder and semantic segmentation features $\psi$ are provided as input to the interaction module, where the interaction between the agents (\textit{social} interaction) and with the environment (\textit{environmental} interaction) are discovered. The encoded interaction features $X_i$ are concatenated with the localization field ($L_{t-1}$, as shown in Fig.\ref{fig:1}) that is computed at the previous time step through the future motion decoder. Note that at the first future time step $T_{obs+1}$, we use the image-like tensor as $L_{obs}$, which is processed from the last observation. We send their concatenation $[X_i,L_{t-1}]$ to the future motion decoder and produce composite fields that are used to decode the locations of agents at all future time steps. 

\subsection{Composite Fields}
The output of the network consists of two types of fields: \textit{localization} and \textit{association} fields. The \textit{localization} field $L_t$ is used to find the locations of all agents at each future time step. In order to identify each agent's trajectory, the locations at different time steps should be associated across time. We use \textit{association} field $A_t$ to associate the agent's location $P_i^t$ at time $t$ with its past location $P_i^{t-1}$ at time $t-1$. All fields generated by the network have a dimension of $64\times64$.

\subsubsection{Localization field} 
\begin{figure*}[t]

    \centering
    \includegraphics[width=0.034\textwidth]{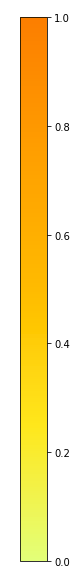}
    \subfigure[]{\includegraphics[width=0.3\textwidth]{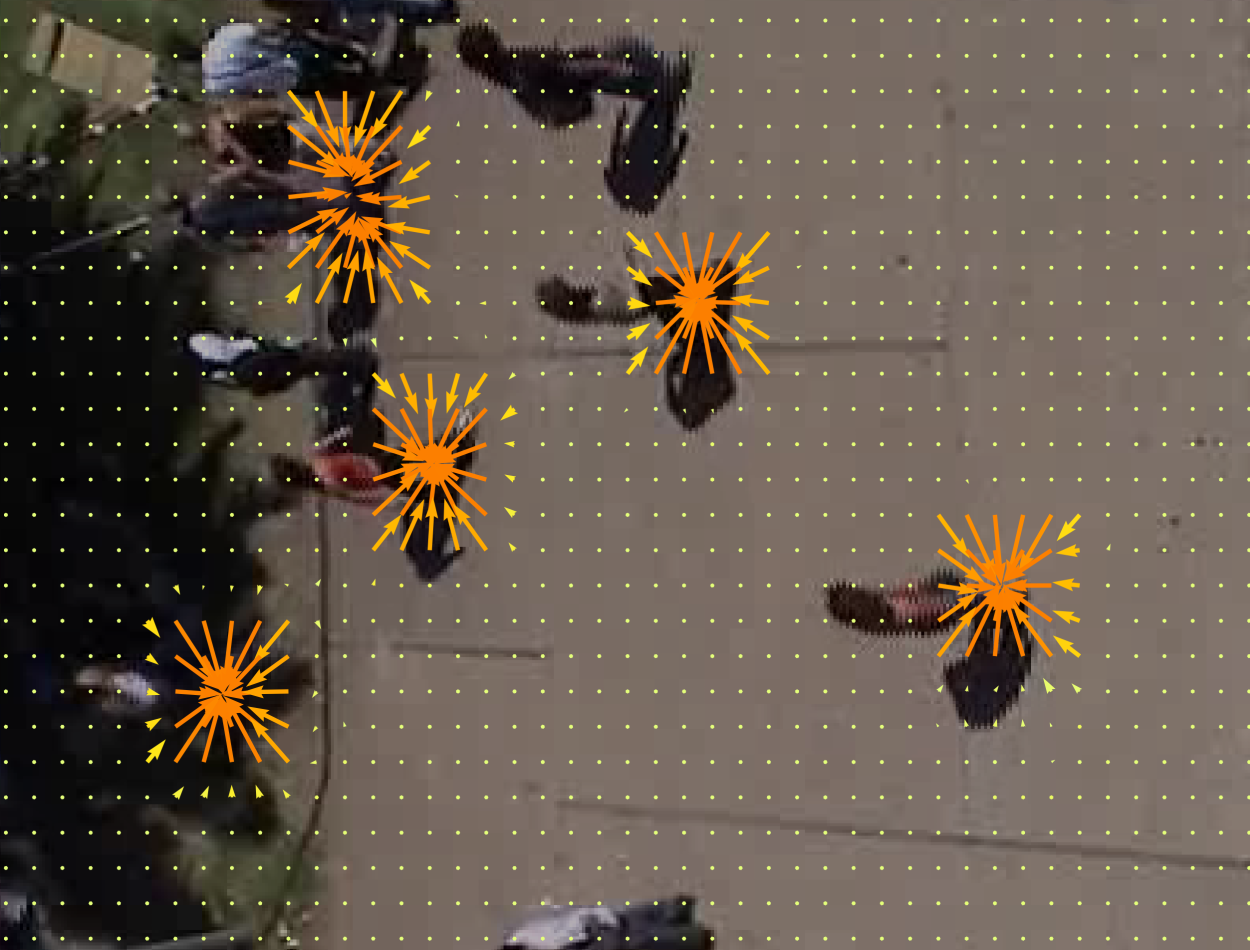}} 
    \subfigure[]{\includegraphics[width=0.3\textwidth]{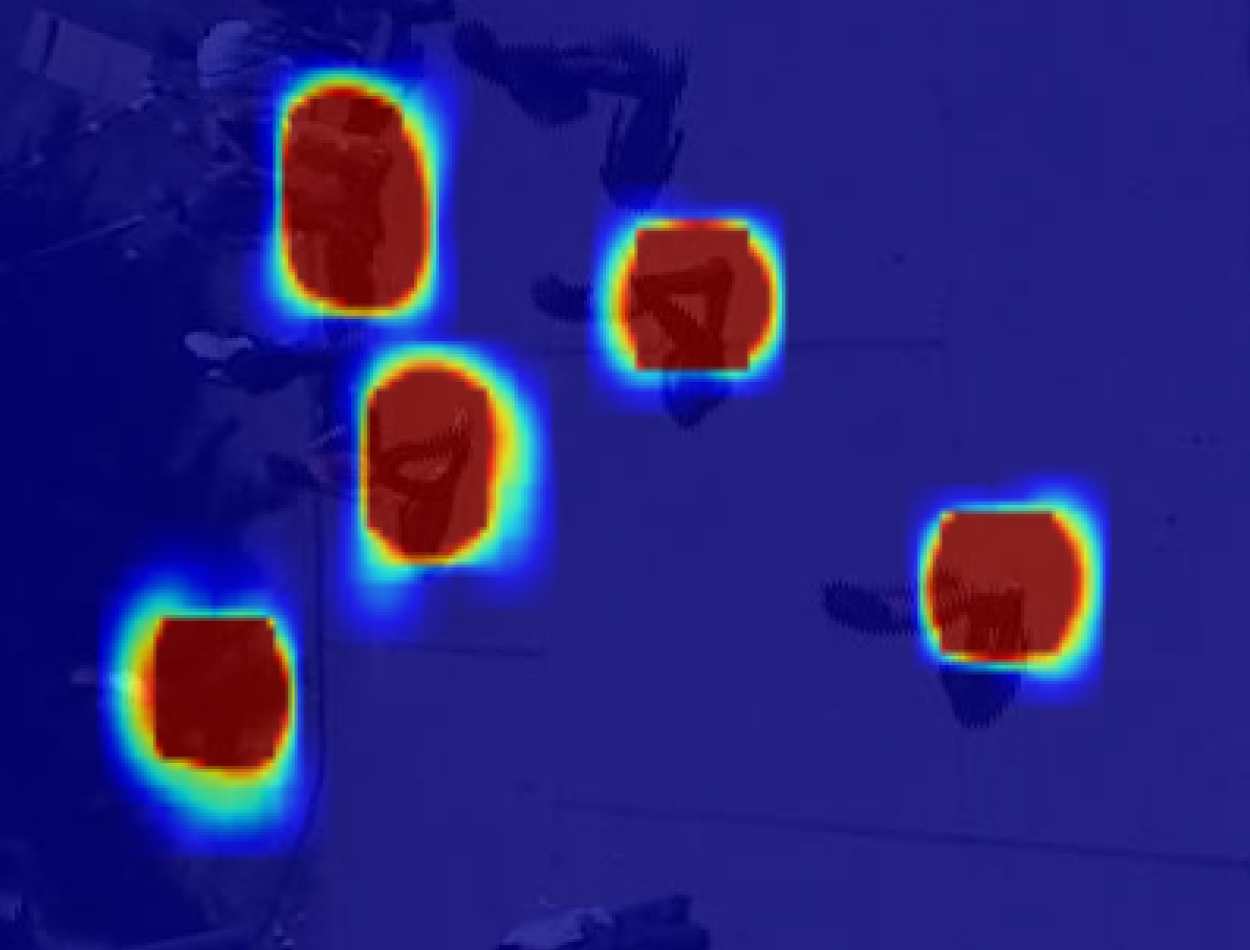}} 
    \subfigure[]{\includegraphics[width=0.3\textwidth]{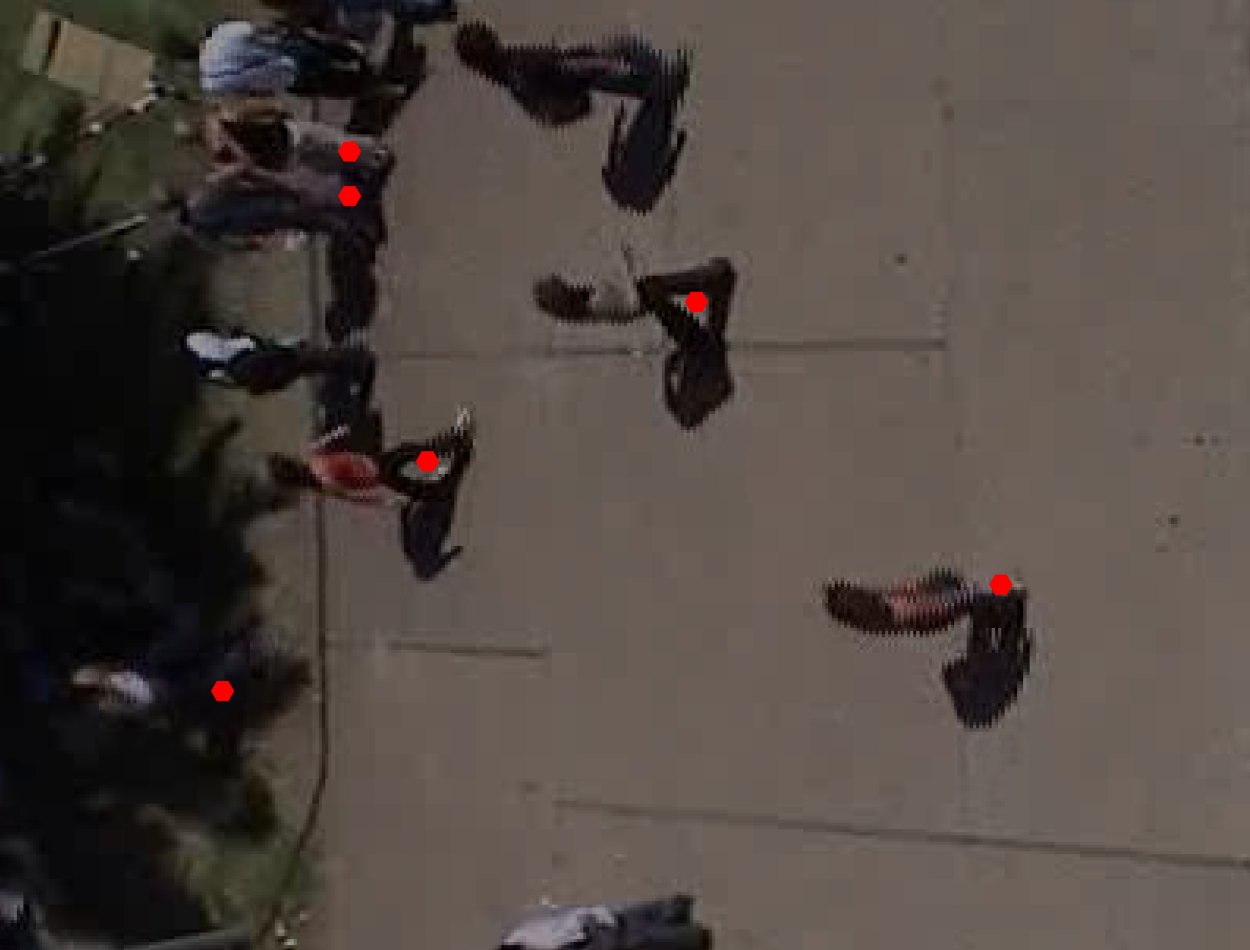}}
    \caption{ \textit{localization} composite fields shown by $(a)$ at an instance from \textit{univ} when used as the test set. These fields are used to predict the locations of all agents at the next time step. Each point represents 3 values $(x,y,p)$. $(x,y)$ is represented by the vector at each point and $p$ is represented by the color of the vector. The matrix \textbf{H} which is formed by weighting the composite fields is shown in figure (b). Peaks detected in H, which are the detected locations of pedestrians are shown in (c). }
    \label{fig:localization}
\end{figure*}

At each spatial location $(i,j)$ in the field, the network predicts 3 parameters $(x_{ij}, y_{ij}, p_{ij})$ representing directional offset $(x_{ij}, y_{ij})$ and its confidence $p_{ij}$. If $(i,j)$ is within a certain threshold Manhattan distance $d_0$ away from an agent location, each triplet $(x_{ij}, y_{ij}, p_{ij})$ would represent a prediction of the position of this agent. 
When $(i,j)$ is within distance $d_0$ away from more than one agents, $(x_{ij}, y_{ij}, p_{ij})$ would represent a prediction of the position of the nearest agent.
If a point $(i,j)$ on the field is not in the vicinity of any agent, then $x_{ij}= 0 , y_{ij}=0, p_{ij} = 0$. Thus, each agent's location is predicted by multiple points $(i,j)$ in its vicinity defined by $d_0$. Fig.~\ref{fig:localization} shows an example of localization field generated by our network. The final location is an ensemble of the predictions. To predict these locations, we create map, $\textbf{H}$ which accumulates all the predictions. For each spatial location in the field, we add a Gaussian contribution in $\textbf{H}$, such that, 
\begin{equation}
\textbf{H} = \sum_{i,j}p_{ij}\mathcal{N}\Big(\hat{\mu},\bm{\Sigma} \Big),
\label{eq:1}
\end{equation}
where the mean $\mu=(i,j)$ is shifted using the directional offset $(x_{ij}, y_{ij})$ predicted by the localization field as $\hat{\mu} = (i + x_{ij}, j + y_{ij})$, and $\bm{\Sigma}$ represents a constant covariance matrix. 

Although the future motion decoder outputs localization fields $L$ of dimensions $64\times64$ for computational efficiency, it is up-scaled to $256\times256$ while finding locations of agents in Eqn.~\ref{eq:1}. The peaks detected on \textbf{H} are the predicted future locations for all agents. We use thresholding followed by 2D non-maximum suppression to find peaks in $\textbf{H}$.
\begin{figure}[H]
\centering
\includegraphics[width=1 \textwidth]{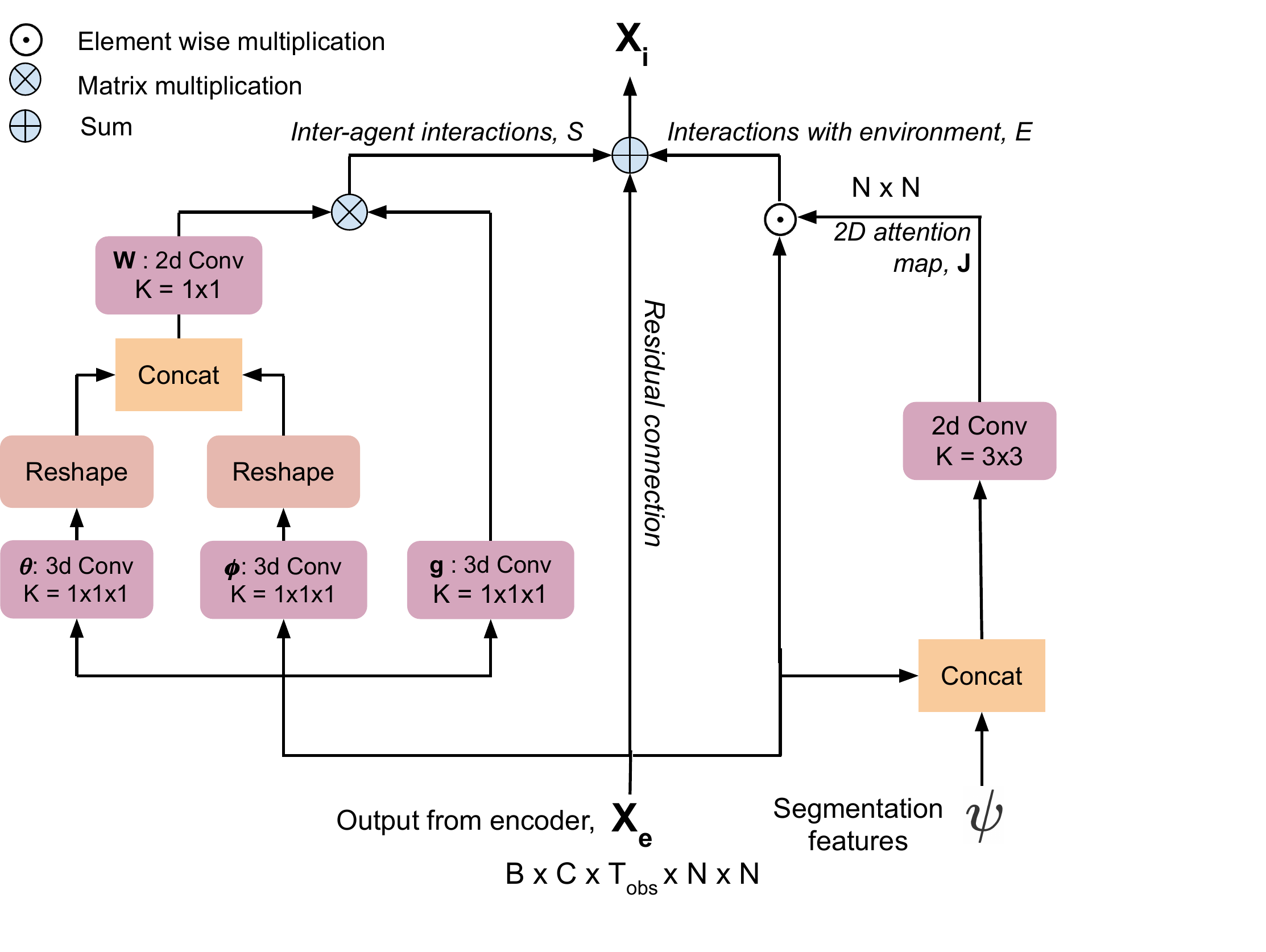}
\caption{Interaction module: the left branch encodes the Social interaction, S and the right branch encodes the environmental interactions, E. The central branch is the residual connection.}
\label{fig:NL}
\end{figure}
\begin{figure*}[t]

    \centering
    \includegraphics[width=0.034\textwidth]{Figures/colormap.png}
    \subfigure[]{\includegraphics[width=0.3\textwidth]{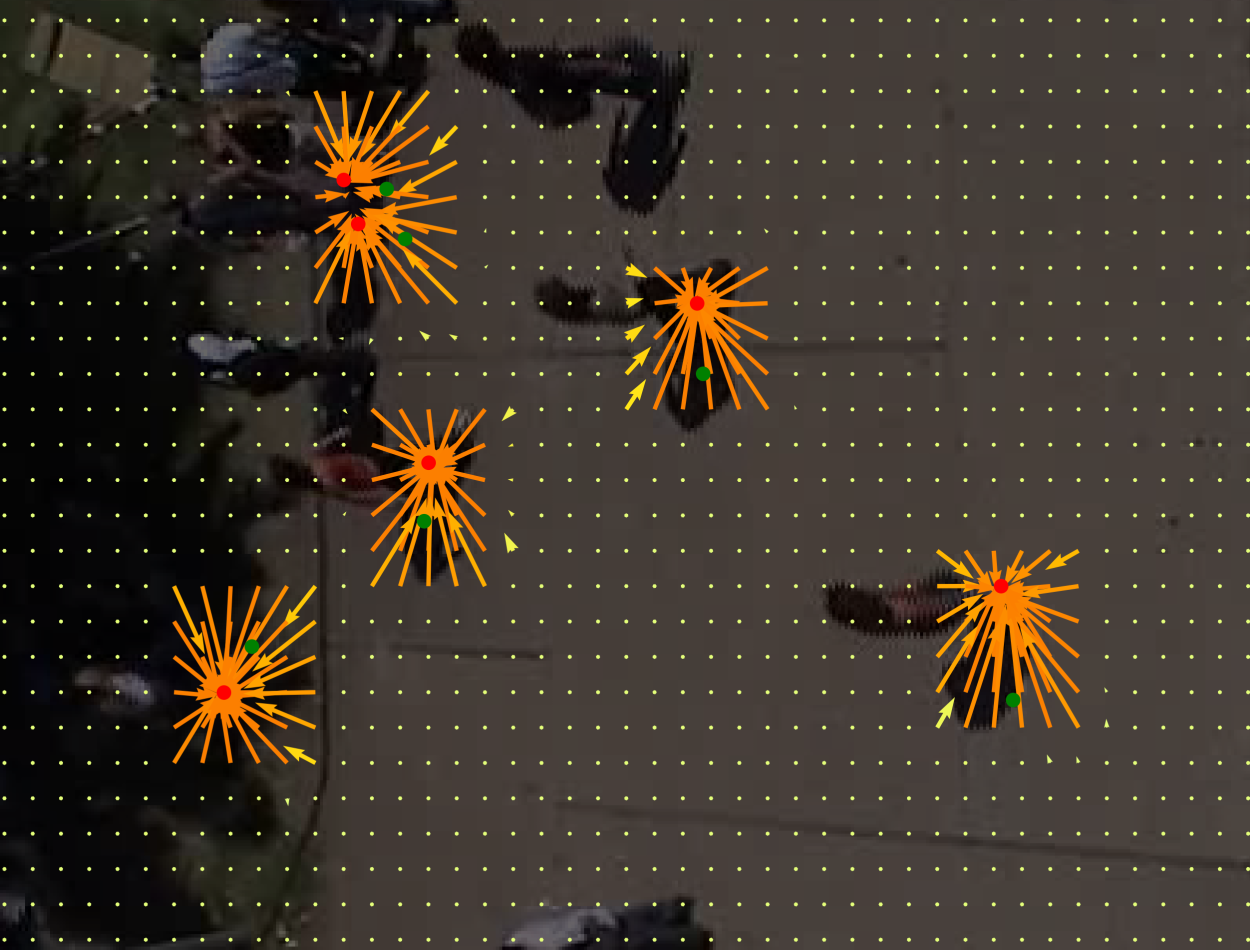}} 
    \subfigure[]{\includegraphics[width=0.3\textwidth]{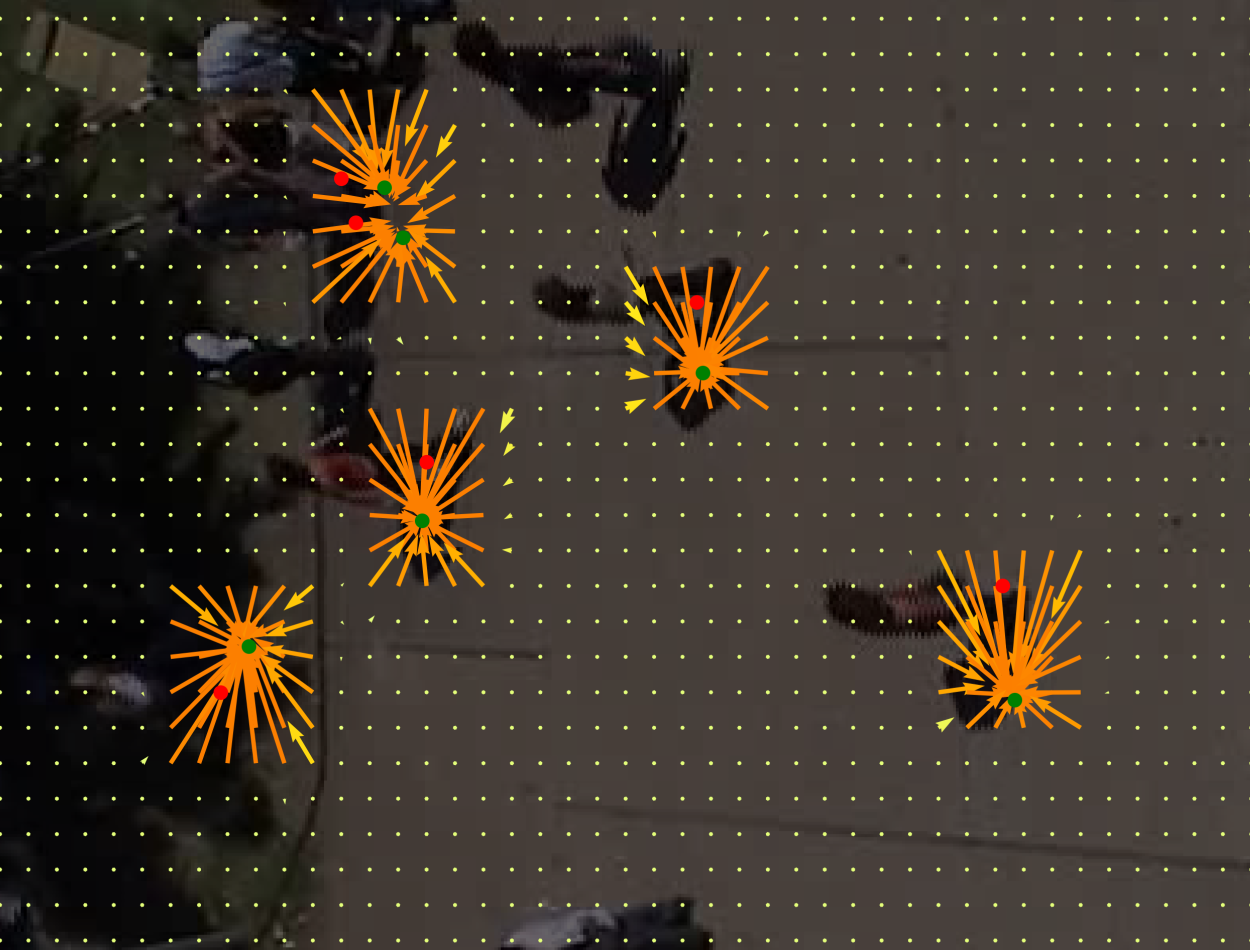}} 
    \subfigure[]{\includegraphics[width=0.3\textwidth]{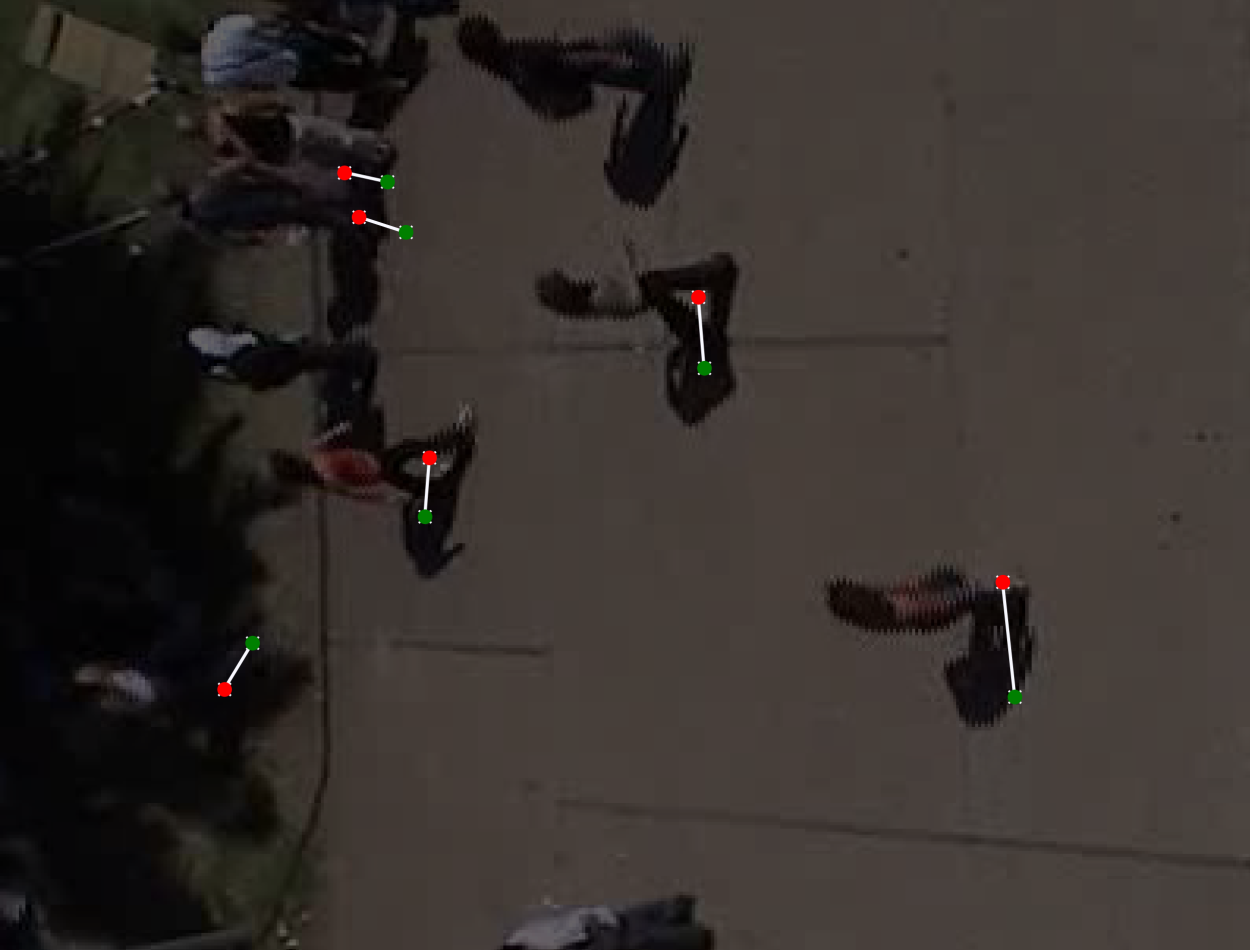}}
    \caption{Subplots $(a), (b)$ show  \textit{association} fields at an instance from \textit{univ} when used as the test set. The image is cropped for better viewing. These fields are used to associate the location of a pedestrian at time $t-1$ (shown in green) with its location at $t$ (shown in red). (c) shows the association between pedestrian locations at $t$ and $t-1$. The background image shows the location of pedestrians at time $t$. }
    \label{fig:association}
\end{figure*}
\subsubsection{Association fields} 
At each spatial location $(i,j)$ of association field, the network predicts 5 parameters $(x^{t-1}_{ij},y^{t-1}_{ij},x^{t}_{ij},y^{t}_{ij}, p_{ij})$. If $(i,j)$ is within a certain threshold Manhattan distance away from an agent's location, then the directional vector $(x^{t-1}_{ij},y^{t-1}_{ij})$ originating at $(i,j)$ would point to the location of the agent at time $t-1$. Similarly the directional vector $(x^{t}_{ij},y^{t}_{ij})$ originating at $(i,j)$ would point to the location of the agent at time $t$. Similar to the localization fields, if the point $(i,j)$ on the field is not in the vicinity of any agent, then all the parameters will be equal to $0$.  During testing, say, we need to associate the location of a pedestrian W at time $t+1$(from a set of candidate locations which were already found using the localization maps), given we already know W's location at time $t$ as $(u,v)$. We threshold the association fields with a threshold on $p_{ij}$. Then we find the directional vector  $(x^{t-1}_{ij},y^{t-1}_{ij})$ which gives the best prediction of W's position at $t$, $(u,v)$. The corresponding $(x^{t}_{ij},y^{t}_{ij})$ would point towards W's estimated position at $t+1$. From the list of candidate points, we find the point nearest to this estimate.
\subsection{Semantic Context}
Future trajectory of road agents depends on their interactions with the environment. For example, agents will change their trajectory to avoid collisions not just with other agents but also obstacles in the scene. Some types of agents may also be more likely to traverse in certain areas of the scene than others. Pedestrians are more likely to travel on the side-walk as opposed to an area occupied by grass. Such contextual information should be considered for more accurate and natural motion prediction. Therefore, we model environmental interactions of road agents using semantic features. To extract such features, we annotate a segmentation map for each scene in the ETH, UCY, and SDD datasets. We define five classes for annotation; walkable area, area covered by grass / bushes / trees, drivable areas, non-drivable areas, and sidewalks. Each pixel can belong to more than one of these classes. We use a small network with 4 convolutional layers to extract contextual features from this annotated map during training. These extracted features are given as input to the interaction module to encode the agents' interaction with the environment.

\subsection{Interaction Module}
We note the local nature of the convolutional and recurrent operations. Therefore they are able to only capture interactions occurring in the local neighborhood either in space or in time.

The recently introduced non-local block \cite{wang2018non} overcomes such locality issues. 
Since the output of the non-local block at a certain location is the linear weighted combination of features at all other locations, they are well suited to capture interactions in space-time domains. We redesign their non-local block by taking its original advantage of interaction modeling capability  (\textit{social} interaction between agents) while additionally capturing \textit{environmental} interactions (interaction between agents and environment) across different spatial and temporal locations. 

Fig.~\ref{fig:NL} shows the proposed non-local block. The semantic context of the scene guides the interaction module
to consider the environmental constraints towards the agents’ potential motion.
Section \ref{results} shows the effectiveness of our modification compared to the vanilla non-local block. The new modified block is able to capture both the inter-agent interactions and environmental interactions. The input to the non-local block is $X_e$ and $\psi$. $X_e$ is obtained from the encoder and $\psi$ is the semantic features extracted from segmentation maps. We take all the hidden states of the last Conv-LSTM of past trajectory encoder at each time step and concatenate them to get $X_e$ of size $B \times C \times T_{obs} \times N \times N$, where $B \times C \times N \times N$ is the size of the feature at each time step, $C$ is the number of channels for the feature, and $B$ is the batch size. The interaction module has 3 branches,
\begin{equation}
I(\psi,X_e) = S(X_e)+E(X_e,\psi)+X_e.
\end{equation}
The first 2 terms, the functions $S$ and $E$ are respectively \textit{social} interaction and \textit{environmental} interaction, which corresponds to the left and right branch in Fig.~\ref{fig:NL}. The last term represents the residual connection. 
$S$ models non-local module as self-attention following~\cite{wang2018non}. For an input feature $z$ this is defined as,
\begin{equation}
q_i = \frac{1}{D(z)}\sum_{\forall j} f(z_i,z_j) g(z_j),
\label{NL_e}
\end{equation}
\begin{table*}[t]
\centering\normalsize
\scalebox{1}{
\begin{tabular}{ l|lllll|l}
\hline
&ETH&~Hotel&~Univ&~Zara1&~Zara2&Avg\\\hline\hline
\textit{State of the art}&&&&&\\
Linear&0.143 / 0.298&~0.137 / 0.261&~0.099 / 0.197&~0.141 / 0.264&~0.144 / 0.268&0.133 / 0.257\\
S-LSTM \cite{alahi2016social}&0.195 / 0.366&~0.076 / 0.125&~0.196 / 0.235 &~0.079 / 0.109&~0.072 / 0.120&0.124 / 0.169\\ 
SS-LSTM \cite{xue2018ss}&0.095 / 0.235&~0.070 / 0.123&~0.081 / 0.131&~0.050 / 0.084&~0.054 / 0.091&0.070 / 0.133\\
SGAN-P \cite{gupta2018social} &0.091 / 0.178&~0.052 / 0.094&~0.112 / 0.215&~0.064 / 0.130&~0.059 / 0.115& 0.075 / 0.146\\
Gated-RN \cite{choi2019looking}&0.052 / 0.100&~0.018 / 0.033&~0.064 / 0.127&~0.044 / 0.086&~\textbf{0.030} / \textbf{0.059}&0.044 / 0.086\\

\hline
\textit{Ours}&&&&&\\
ED&0.051 / 0.085 &~0.024 / 0.039&~0.079 / 0.156&~0.058 / 0.121&~0.056 / 0.113&0.054 / 0.103\\
ED+F&0.043 / 0.077&~0.018 / 0.029&~0.069 / 0.141&~0.043 / 0.090&~0.049 / 0.100&0.045 / 0.088 \\
ED+F+$I_1$&0.042 / 0.072&~0.019 / 0.034&~0.064 / 0.128&~0.040 / 0.081&~0.050 / 0.102&0.043 / 0.083\\
ED+F+$I_2$&0.040 / 0.075&~0.020 / 0.038&~0.063 / 0.131&~0.040 / 0.084&~0.045 / 0.095&0.042 / 0.084\\
ED+F+$I_3$ &0.038 / 0.067&~\textbf{0.016} / \textbf{0.028}&~0.060 / 0.122&~0.040 / 0.080&~0.048 / 0.097&0.040 / 0.078 \\ 
ED+F+$I_4$&\textbf{0.036} / \textbf{0.064}&~0.018 / 0.031 &~\textbf{0.059} / \textbf{0.120}&~\textbf{0.038} / \textbf{0.078}&~0.046 / 0.094&\textbf{0.039} / \textbf{0.077} \\ \hline
\end{tabular}}
\caption{Two error metrics ADE / FDE at $T_{pred}$ = 12 time-steps in normalized pixels on ETH and UCY datasets. I1 uses \textit{concatenation} for interaction modelling, I2 uses a small \textit{3D conv net}, I3 uses a \textit{non-local} interaction module and I4 uses the modified non-local block which also models physical interactions.}
\label{eth_ucy}
\end{table*}

\noindent
where $z_i$ is the index of an output position (in space-time) and $j$ is the index that enumerates all possible locations. $f$ is a pair-wise function that computes the relationship between $z_i$ and $z_j$. $g(z_j)$ calculates the weight corresponding to weighting of the self-attention and $D(z)$ is the normalizing factor. For our case, we use a concatenation version of f:
\begin{equation}
    f(X_{ei},X_{ej}) = ReLU(W[\theta(X_{ei}), \theta(X_{ej})]),
\end{equation}
where $\theta, \phi, g$ are convolutional layers as shown in Fig.~\ref{fig:NL} and $[\cdot,\cdot]$ represents the concatenation operation. The matrix multiplication in Fig.\ref{fig:NL} is responsible for the multiplication of f,g and the successive summation over $j$ in Eqn.\ref{NL_e}.\\
$E$ is obtained by element-wise multiplication of $B \times C \times T_{obs} \times N \times N$ shaped $X_e$ with a $N \times N$ shaped attention heat-map, $J$ produced using the segmentation features. 
\begin{equation}
  E(X_e, \psi) = X_e \odot J
 \end{equation}
The element-wise multiplication allows the network to suppress a certain region of the input feature maps, which may not be useful for trajectory prediction (\textit{e.g.}, non-walkable areas).



\section{Experiments}



\begin{table}[h!]
\centering\small
\scalebox{1}{
\setlength\tabcolsep{1.6pt}
\begin{tabular}{ l|l|l|l|l|l}
\hline
\multicolumn{5}{c|}{\textit{State of the art}}& 
\multicolumn{1}{c}{\textit{Ours}}\\\hline
&Linear&S-LSTM&DESIRE&Gated-RN&ED+I4\\ \hline
1.0 \textit{sec}&- / 2.58&1.93 / 3.38&- / 2.00&1.71 / 2.23&\textbf{1.57} / \textbf{2.06}  \\
2.0 \textit{sec}&- / 5.37&3.24 / 5.33&- / 4.41&2.57 / 3.95&\textbf{2.48} / \textbf{3.91}  \\
3.0 \textit{sec}&- / 8.74&4.89 / 9.58&- / 7.18&3.52 / 6.13&\textbf{3.47} / \textbf{6.08}  \\
4.0 \textit{sec} &- / 12.54&6.97 / 14.57&- / 10.23&4.60 / 8.79&\textbf{4.53} / \textbf{8.43}  \\\hline

\end{tabular}}

\caption{Comparison of our method with the state-of-the art on the SDD dataset. The error metrics ADE and FDE (in pixels) at $T_{pred}$ = 10 time steps are shown at a scale of 1/5 in accordance to the experimental settings of \cite{lee2017desire}.}
\label{sdd}
\end{table}

\subsection{Datasets}
We use three publicly available benchmark datasets (ETH, UCY, and SDD) for our experiments. The ETH~\cite{pellegrini2009you} and UCY~\cite{lerner2007crowds} datasets contain top view videos of pedestrians walking at public locations. Combined, these datasets contain 5 scenes {eth, hotel, ucy, zara1, zara2} that are captured using a stationary camera. The SDD dataset~\cite{robicquet2016learning} contains 60 unique top view videos taken using a drone at a university campus. The dataset consists of pedestrians, cyclists, and cars.

These datasets contain naturalistic interactions between road agents and between agents and the environment. Both datasets have annotated labels of the agents' locations in world coordinates. In our experiments, we convert all agents' locations to pixel locations in a $256\times256$ sized image space. For ETH / UCY datasets, we use a leave-one-out cross validation policy, where we train on $4$ splits and test on the remaining $5^{th}$ split. For SDD, we follow the same settings used by~\cite{lee2017desire}.
\subsection{Implementation details}
\label{implementation_details}
We used Nvidia V100 and P100 GPUs for all experiments. Our network is implemented in PyTorch and  trained from scratch using the He Normal~\cite{he2015delving} initialization. For optimization, we use Adam~\cite{kingma2014adam} with a mini-batch size of $20$ and initial learning rate of $5\times 10^{-5}$ with step decay. We train our models for 100 epochs. Each Conv-LSTM layer is followed by a normalization layer in order to facilitate convergence of the network (these layers are not shown in Fig.\ref{fig:1}). We employ data augmentation to reduce over fitting, whereby training videos are randomly flipped horizontally / vertically and rotated in multiples of 90\degree  about the center. Our network uses a Mean square error (MSE) to learn composite fields.

For ETH / UCY datasets, our method observes the trajectory of agents for 8 time-steps ($T_{obs} = 8$, corresponding to 3.2 sec) and predicts the  trajectory for the next 12 time-steps ($T_{pred}= 12$, corresponding to 4.8 sec). For the SDD dataset, we use $T_{obs} = 8,\ T_{pred}= 10$ (corresponding to 3.2 sec and 4.0 sec, respectively) to keep experiments comparable with~\cite{lee2017desire}. During training, we prepared ground truth \textit{localization} and \textit{association} fields which are learned by our network by regression. The confidence parameter of both the \textit{association} and \textit{localization} field, $p_{ij}=1$ for all ground truth fields. We use the MSE loss for regression. 

The input to the past motion encoder at time $t (\forall \in [1, T_{obs}])$ is a binary map $M_t$ of size $(256,256)$. If $(x_i^{t},y_i^{t})$ is the location of an agent at time-step $t$, then all points in $M_t$ within a $10$ Manhattan distance from $(x_i^{t},y_i^{t})$ would have value $1$. All points in $M_t$ which do not satisfy this condition will have value $0$. 

\subsection{Baselines and Evaluation}

We compare our work with the state-of-the-art models in the literature. We use Linear, Social-LSTM~\cite{alahi2016social}, SS-LSTM~\cite{xue2018ss}, and Gated-RN~\cite{choi2019looking} for comparison on ETH / UCY dataset. For the SDD dataset, we use Linear, Social-LSTM, Gated-RN, and DESIRE~\cite{lee2017desire}. Linear is a linear regressor that estimates future trajectory by minimizing the mean square error. 

We also show ablative studies of our model to emphasize the importance of each module proposed in our framework. The baseline \textit{ED} is the vanilla encoder-decoder version without any interaction module. This baseline also does not weight the multiple predictions in the \textit{localization} fields to generate a single prediction -- we find the prediction with the highest confidence $p_{ij}$ without using Eqn.~\ref{eq:1}. 
 In the baseline \textit{ED+F}, we use the aggregation of the multiple predictions on the \textit{localization} fields using Eqn.~\ref{eq:1}. Its variations  \textit{ED+F+$I_1$}, \textit{ED+F+$I_2$}, \textit{ED+F+$I_3$} 
use the hidden states of the last Conv-LSTM layer of the past motion encoder in different ways. 
The suffix \textit{$I_1$} concatenates all the hidden-states over the past time steps [$1,...,T_{obs}$], \textit{$I_2$} is a small 3D convolutional network to process these temporal features, and  \textit{$I_3$} is the original non-local interaction module introduced in~\cite{wang2018non}. Finally, our best model with the proposed interaction module \textit{ED+F+$I_4$} advances the non-local module to factor in environmental interactions,  as shown in Fig.\ref{fig:NL}.

Following the standard evaluation metrics, experiments are evaluated using average displacement error (ADE) and final displacement error (FDE). ADE is defined as the mean of L2 distances between the prediction and the ground truth for all time-steps. FDE is defined as the L2 distance between the predicted location at the last time step and the corresponding ground truth. 
\section{Results}
\label{results}

\subsection{Quantitative results}
Table~\ref{eth_ucy} shows a comparison of our approach against state-of-the-art methods as well as our ablation experiments for the ETH and UCY datasets. As expected, \textit{Linear} which is a linear regressor, has the worst performance. S-LSTM shows improved performance due to its use of social pooling. By adding high-level image features, SS-LSTM reaches lower error rates than S-LSTM. Gated-RN shows further improvement due to their pair-wise interaction encoding.

Our first baseline model, \textit{ED} performs better than SS-LSTM. The next model, \textit{ED+F} which incorporates a weighted aggregation of \textit{localization} fields, generates a significant performance boost across all splits, demonstrating the efficacy of our ensemble model. Note that this baseline model is already comparable to all state-of-the-art methods. Through the interaction modeling ablations, we notice that \textit{$I_1$} shows improvement over the baseline \textit{ED+F}, suggesting the use of additional temporal features for the decoder. Further improvement of \textit{$I_2$} suggests that 3D convolutions are more suitable in modeling interactions as compared to naive concatenation of \textit{$I_1$}. \textit{$I_3$}'s usage of the non-local block allows the system to capture spatio-temporal features
, thus showing improvement over \textit{$I_2$}. Using the semantic context extracted from the segmentation features makes the network being aware of physical boundaries and constraints, which results in further improvement of \textbf{$I_4$} for all splits except for hotel. For hotel, we observe that most of the area in the video frames is walkable
, meaning that the segmentation features may not provide any additional context. Given the results in Table~\ref{eth_ucy}, we show that our best model \textit{ED+F+$I_4$} generally outperforms the state-of-the-art methods on most of the splits including the total ADE/FDE. We also compare our method \textit{ED+F+$I_4$} with the existing works using the SDD dataset. In Table\ref{sdd}, our approach achieves performance higher than these state-of-the-art methods over all time steps. The evaluation on ETH, UCY, and SDD datasets demonstrates that the efficacy of our single-shot framework as well as the proposed interaction module. 


\begin{table}[t!]
\centering\small
\scalebox{1.0}{
\setlength\tabcolsep{1.6pt}
\begin{tabular}{l|c|c|c|l} \hline
\multicolumn{5}{c}{average run-time (in \textit{sec}) / speed-up}\\
\hline
No. agents  & {S-LSTM}\cite{alahi2016social}& {S-GAN-P}\cite{gupta2018social}  & {Gated-RN}\cite{choi2019looking}& {Ours} \\ \hline\hline
1(min)  & ~2.61 / 1x  & {0.15} / {17x} &\textbf{0.12} / \textbf{21x} &0.32 / 8x\\ 
4.5(avg)& 11.69 / 1x & 0.67 / 17x &0.54 / 21x & \textbf{0.32} / \textbf{36x} \\
21(max) & 54.80 / 1x&3.15 / 17x &2.52 / 21x& \textbf{0.32} / \textbf{171x} \\\hline
\end{tabular}
}
\caption{Comparison of the computational performance of the proposed approach with the state-of-the-art methods. We report average run-time / speed-up with respect to S-LSTM~\cite{alahi2016social} evaluated using the ETH, UCY datasets. Note that the average number of road agents in these datasets is 4.5 per scene, and we achieve 36x speed-up. The minimum and maximum number of agents in a scene in these datasets is 1 and 21 respectively.}
\label{speed}
\end{table}

\noindent
\subsection{Speed and Run time}
The computational time of such systems is one of the critical aspects for their practical use in real world applications such as autonomous navigation for robotics and driving scenarios. Table\ref{speed} shows the comparison of our method against S-LSTM~\cite{alahi2016social}, S-GAN-P~\cite{gupta2018social}, and Gated-RN~\cite{choi2019looking}. 
\begin{figure*}[!t]
\centering
\includegraphics[width=1 \textwidth]{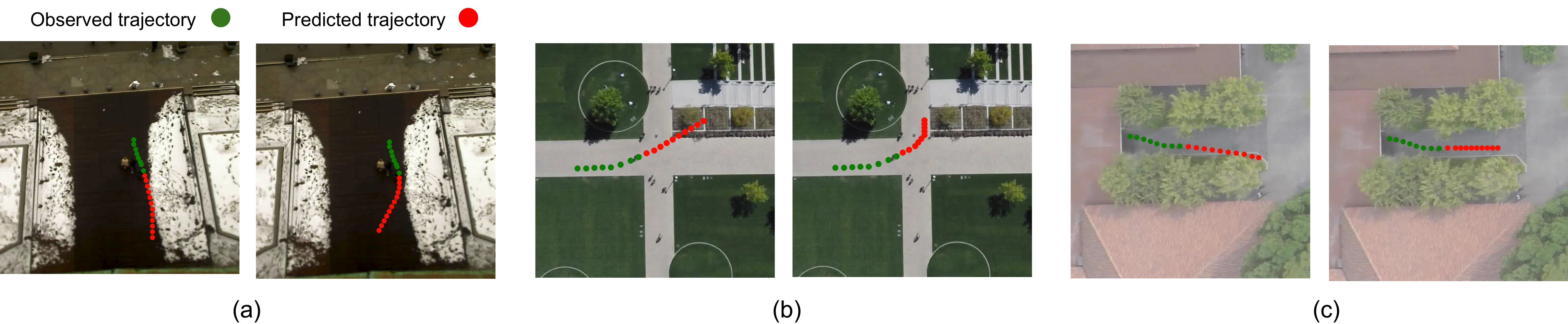}
\caption{Qualitative results comparing  \textit{ED+F+$I_3$} (left subplot) and  \textit{ED+F+$I_4$} (right subplot). We observe that using segmentation information (\textit{ED+F+$I_4$}) allows future trajectory to conform to the physical environmental conditions. By using segmentation features, the predicted future trajectory of the pedestrian \textit{(a)}  avoids the snow accumulation, \textit{(b)} stays on the paved walkway, and \textit{(c)} avoids entering the shrub.}
\label{seg_improv}
\label{inter}
\end{figure*}
\begin{figure}[!tbp]
\centering
\includegraphics[width=1 \textwidth]{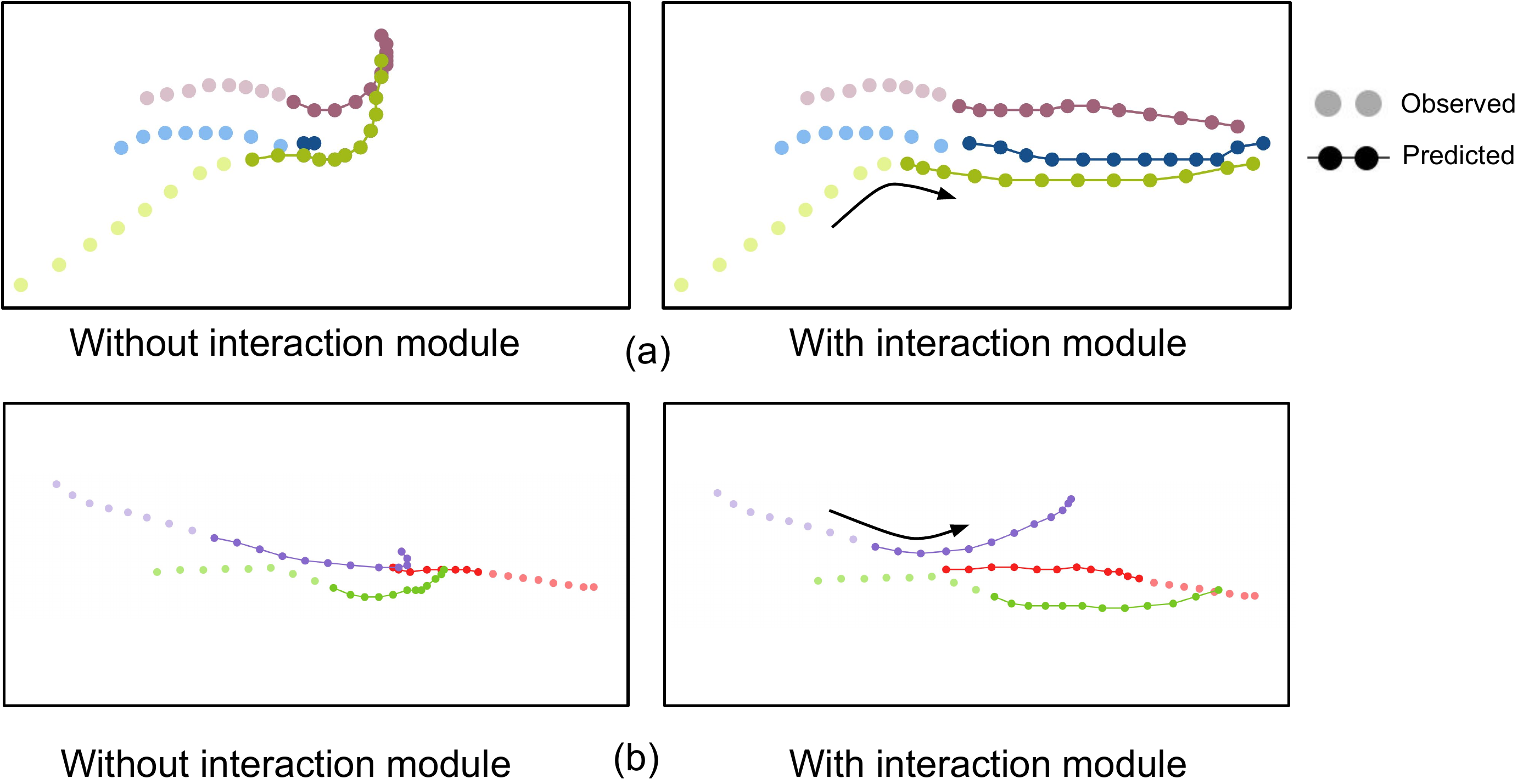}
\caption{Figure illustrating \textit{social} interaction improvement as a result of the interaction module \textit{$I_3$} (right) as compared to \textit{$I_1$} (left) for samples in (a) zara1  and (b) univ, respectively. The lighter colored markers represent the observed trajectory while the darker colored markers connected with a line represent the predicted trajectory. (a), The pedestrian marked by the green colored marker changes direction to avoid collision (emphasized by the black arrow). (b) The pedestrian marked by the purple colored marker changes direction to avoid collision with the red colored pedestrian (emphasized by the black arrow).}
\label{inter_improv}
\end{figure}
Although the time complexity of existing methods linearly increases as O(n) where n is the number of agents, the proposed approach always runs in constant time complexity O(1) due to its structure, specifically designed for single-shot prediction. In practice, our model is able to run 171x faster than S-LSTM\cmmnt{, 9x faster than S-GAN,} and 7x faster than Gated-RN with 21 observed agents, while achieving higher performance as reported in Table~\ref{eth_ucy} and \ref{sdd}.

\subsection{Qualitative results}
Fig. \ref{seg_improv} illustrates the impact of the proposed interaction module \textit{$I_4$} against the non-local block \textit{$I_3$}. The examples are collected from ETH and SDD datasets, highlighting qualitative improvement due to the usage of semantic context for environmental interactions. Subsequently, we show the efficacy of the proposed interaction module for inter-agent interactions. Fig. \ref{inter_improv} compares the results of future trajectory prediction with and without the interaction module. The examples are drawn from the UCY dataset, and they clearly show that the future motions are being more natural with the interaction module considering potential collisions between agents.

\section{Conclusions}

In this paper, we considered the problem of future trajectory forecast using composite fields for single-shot prediction of all agents' future trajectories.  We demonstrated the efficacy of the proposed models with experimental evaluations on ETH, UCY, and SDD datasets.  The results showed that our novel interaction module improves performance by capturing inter-agent and environmental interactions in both spatial and temporal domains.  Importantly, unlike previous methods in trajectory forecast, the proposed network is highly efficient in its computation and  runs in constant time with respect to any number of agents in the scene.   In the future, we plan to extend this work to egocentric video inputs obtained from a moving platform and incorporate additional scene contexts to capture participant interactions with the environment in various autonomous mobility applications.






\bibliographystyle{ieeetr}
\bibliography{main}

\end{document}